\documentclass[11pt,letterpaper]{article}
\usepackage[letterpaper]{geometry}
\usepackage{conll2018}
\usepackage{times}
\usepackage{latexsym}
\usepackage{amsmath}
\usepackage{multirow}
\usepackage{url}
\usepackage[T1]{fontenc}
\usepackage{graphicx}
\usepackage{booktabs}
\usepackage{array}
\usepackage[inline]{enumitem}
\usepackage{amsmath}
\usepackage{color}
\usepackage{graphicx}
\usepackage{microtype}
\usepackage[smallscripts]{moresize}

\aclfinalcopy 

\makeatletter
\newcommand{\@emptybiblabel}[1]{}
\makeatother

\newcommand{\caret}{^}

\title{Modeling Composite Labels for Neural Morphological Tagging}

\author{Alexander Tkachenko \\
  Institute of Computer Science \\ University of Tartu \\ Tartu, Estonia \\
  {\tt aleksandr.tkatsenko@ut.ee} \\\And
  Kairit Sirts \\
  Institute of Computer Science \\ University of Tartu \\ Tartu, Estonia \\
  {\tt kairit.sirts@ut.ee} \\}

\date{}

\begin{document}
\maketitle
\begin{abstract}
Neural morphological tagging has been regarded as an extension to POS tagging task, treating each morphological tag as a monolithic label and ignoring its internal structure.
We propose to view morphological tags as composite labels and explicitly model their internal structure in a neural sequence tagger.
For this, we explore three different neural architectures and compare their performance with both CRF and simple neural multiclass baselines.
We evaluate our models on 49 languages and show that the neural architecture that models the morphological labels as sequences of morphological category values performs significantly better than both baselines establishing state-of-the-art results in morphological tagging for most languages.\footnote{The source code is available at \\ \ssmall{\url{https://github.com/AleksTk/seq-morph-tagger}}}
\end{abstract}

\section{Introduction}
\label{sec:introduction}


The common approach to morphological tagging combines the set of word's morphological features into a single monolithic tag and then, similar to POS tagging, employs multiclass sequence classification models such as CRFs \citep{mueller2013} or recurrent neural networks \citep{labeau2015,heigold2017}.
This approach, however, has a number of limitations.
Firstly, it ignores the intrinsic compositional structure of the labels and treats two labels that differ only in the value of a single morphological category as completely independent; compare for instance labels \textsc{[POS=noun,Case=Nom,Num=Sg]} and \textsc{[POS=noun,Case=Nom,Num=Pl]} that only differ in the value of the \textsc{Num} category.
Secondly, it introduces a data sparsity issue as the less frequent labels can have only few occurrences in the training data.
Thirdly, it excludes the ability to predict labels not present in the training set which can be an issue for languages such as Turkish where the number of morphological tags is theoretically unlimited \citep{yuret2006}.

To address these problems we propose to treat morphological tags as composite labels and explicitly model their internal structure. 
We hypothesise that by doing that, we are able to alleviate the sparsity problems, especially for languages with very large tagsets such as Turkish, Czech or Finnish, and at the same time also improve the accuracy over a baseline using monolithic labels.
We explore three different neural architectures to model the compositionality of morphological labels.
In the first  architecture, we model all morphological categories (including POS tag) as independent multiclass classifiers conditioned on the same contextual word representation.
The second architecture organises these multiclass classifiers into a hierarchy---the POS tag is predicted first and the values of morphological categories are predicted conditioned on the value of the predicted POS.
The third architecture models the label as a sequence of morphological category-value pairs.
All our models share the same neural encoder architecture based on bidirectional LSTMs to construct contextual representations for words \citep{lample2016}.

We evaluate all our models on 49 UD version 2.1 languages.
Experimental results show that our sequential model outperforms other neural counterparts establishing state-of-the-art results in morphological tagging for most languages.
We also confirm that all neural models perform significantly better than a competitive CRF baseline.
In short, our contributions can be summarised as follows:
\begin{enumerate}[label=\arabic*),topsep=0em,noitemsep]
\item We propose to model the compositional internal structure of complex morphological labels for morphological tagging in a neural sequence tagging framework;
\item We explore several neural architectures for modeling the composite morphological labels;
\item We find that tag representation based on the sequence learning model achieves state-of-the art performance on many languages.
\item We present state-of-the-art morphological tagging results on 49 languages on the UDv2.1 corpora.
\end{enumerate}

\section{Related Work}
\label{sec:related}

Most previous work on modeling the internal structure of complex morphological labels has occurred in the context of morphological disambiguation---a task where the goal is to select the correct analysis from a limited set of candidates provided by a morphological analyser. The most common strategy to cope with a large number of complex labels has been to predict all morphological features of a word using several independent classifiers whose predictions are later combined using some scoring mechanism \citep{hajic1998,hajic2000,smith2005,yuret2006,zalmout2017,kirov2017}. 
\citet{inoue2017} combined these classifiers into a multitask neural model sharing the same encoder, and predicted both POS tag and morphological category values given the same contextual representation computed by a bidirectional LSTM. 
They showed that the multitask learning setting outperforms the combination of several independent classifiers on tagging Arabic.
In this paper, we experiment with the same architecture, termed as multiclass multilabel model, on many languages.
Additionally, we extend this approach and explore a hierarchical architecture where morphological features directly depend on the POS tag. 


Another previously adopted approach involves modeling complex morphological labels as sequences of morphological feature values \citep{hakkani2000,schmid2008}. In neural networks, this idea can be implemented with recurrent sequence modeling. Indeed, one of our proposed models generates morphological tags with an LSTM network. Similar idea has been applied for the morphological reinflection task \citep{kann2016,faruqui2016} where the sequential model is used to generate the spellings of inflected forms given the lemma and the morphological label of the desired form. In morphological tagging, however, we generate the morphological labels themselves.

Another direction of research on modeling the structure of complex morphological labels involves structured prediction models~\citep{mueller2013,mueller2015,malaviya2018,lee2011}.
\citet{lee2011} introduced a factor graph model that jointly infers morphological features and syntactic structures.
\citet{mueller2013} proposed a higher-order CRF model which handles large morphological tagsets by decomposing the full label into POS tag and morphology part.
\citet{malaviya2018} proposed a factorial CRF to model pairwise dependencies between individual features within  morphological labels and also between labels over time steps for cross-lingual transfer.
Recently, neural morphological taggers have been compared to the CRF-based approach \citep{heigold2017,yu2017}.
While \citet{heigold2017} found that their neural model with bidirectional LSTM encoder surpasses the CRF baseline, the results of \citet{yu2017} are mixed with the convolutional encoder being slightly better or on par with the CRF but the LSTM encoder being worse than the CRF baseline. 

Most previous work on neural POS and morphological tagging has shared the general idea of using bidirectional LSTM for computing contextual features for words \citep{ling2015,huang2015,labeau2015,ma2016,heigold2017}.
The focus of the previous work has been mostly on modeling the inputs by exploring different character-level representations for words \citep{heigold2016,santos2014,ma2016,inoue2017,ling2015,rei2016}.
We adopt the general encoder architecture from these works, constructing word representations from characters and using another bidirectional LSTM to encode the context vectors. 
In contrast to these previous works, our focus is on modeling the compositional structure of the complex morphological labels. 

The morphologically annotated Universal Dependencies (UD) corpora~\citep{nivre2017} offer a great opportunity for experimenting on many languages. 
Some previous work have reported results on several UD languages \citep{yu2017,heigold2017}. 
Morphological tagging results on many UD languages have been also reported for parsing systems that predict POS and morphological tags as preprocessing \citep{andor2016,straka2016,straka2017}.
Since UD treebanks have been in constant development, these results have been obtained on different UD versions and thus are not necessarily directly comparable. We conduct experiments on all UDv2.1 languages and we aim to provide a baseline for future work in neural morphological tagging.

\section{Neural Models}
\label{sec:models}

\begin{figure*}[]
\centering
\includegraphics[width=0.85\textwidth]{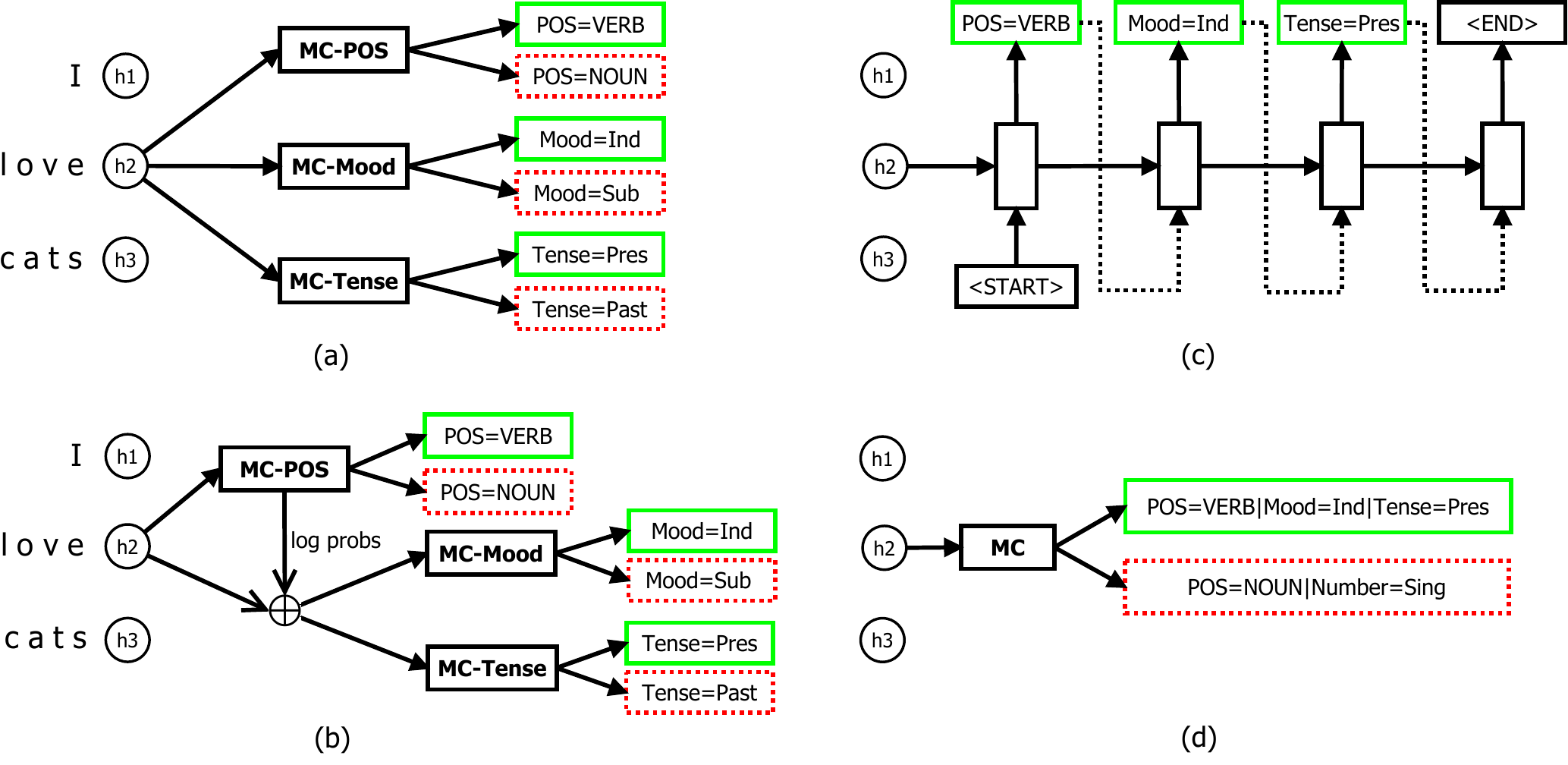}
\caption{Neural architectures for modeling complex morphological labels: a) Multiclass Multilabel model (\textsc{McMl}), b) Hierarchical Multiclass Multilabel model (\textsc{HMcMl}), c) Sequence model (\textsc{Seq}) and d) Multiclass baseline model (\textsc{Mc}). 
Correct labels are shown with a green border, incorrect labels have a red dotted border.}
\label{fig:neural_models}
\end{figure*}

We explore three different neural architectures for modeling morphological labels: multiclass multi\-label model that predicts each category value separately, hierarchical multiclass multilabel model where the values of morphological features depend on the value of the POS, and a sequence model that generates morphological labels as sequences of feature-value pairs.

\subsection{Notation}
Given a sentence $w_1, \dots, w_n$ consisting of $n$ words, we want to predict the sequence $t_1, \dots, t_n$ of morphological labels for that sentence.
Each label $t_i = \{f_{i0}, f_{i1}, \dots, f_{im}\}$ consists of a POS tag ($f_{i0} \equiv \textsc{POS}$) and a sequence of $m$ category values. 
For each word $w_i$, the encoder computes a contextual vector $h_i$, which captures information about the word and its left and right context.

\subsection{Decoder Models}

\paragraph{Multiclass Multilabel model (\textsc{McMl})}
This model formulates the morphological tagging as a multiclass multilabel classification problem. 
For each morphological category, a separate multiclass classifier is trained to predict the value of that category (Figure~\ref{fig:neural_models} (a)).
Because not all categories are always present 
for each POS (e.g., a noun does not have a \textit{tense} category), we extend the morphological label of each word by adding all features that are missing from the annotated label and assign them a special value that marks the category as ``off''.
Formally, the model can be described as:
\begin{equation}
p(t|h)_{\textsc{McMl}} = \prod_{j=0} \caret M p(f_j|h),
\end{equation}
where $M$ is the total number of morphological categories (such as case, number, tense, etc.)
observed in the training corpus.
The probability of each feature value is computed with a softmax function:
\begin{equation*}
p(f_j|h)_{\textsc{McMl}} = \text{softmax}(W_j h + b_j),
\end{equation*}
where $W_j$ and $b_j$ are the parameter matrix and bias vector for the $j$th morphological feature ($ j=0, \dots, M$).
The final morphological label for a word is obtained by concatenating predictions for individual categories while filtering out off-valued categories.

\paragraph{Hierarchical Multiclass Multilabel model (\textsc{HMcMl})}
This is a hierarchical version of the \textsc{McMl} architecture that models the values of morphological categories as directly dependent on the POS tag (Figure~\ref{fig:neural_models} (b)):
\begin{equation}
p(t|h)_{\textsc{HMcMl}} = p(\textsc{pos}|h)\prod_{j=1} \caret M p(f_j|\textsc{pos},h)
\end{equation}
The probability of the POS is computed from the context vector $h$ using the respective parameters:
\begin{equation*}
p(\textsc{pos}|h) = \text{softmax}(W_{\textsc{pos}} h + b_{\textsc{pos}})
\end{equation*}
The POS-dependent context vector $l$ is obtained by concatenating the context vector $h$ with the unnormalised log probabilities of the POS:
\begin{equation*}
l = [h; W_{\textsc{pos}} h + b_{\textsc{pos}}]
\end{equation*}
The probabilities of the morphological features are computed using the POS-dependent context vector:
\begin{equation*}
p(f_j|\textsc{pos}, h) = \text{softmax}(W_j l + b_j) \hspace{2mm} j=1, \dots, M
\end{equation*}

\paragraph{Sequence model (\textsc{Seq})}
The \textsc{Seq} model predicts complex morphological labels as sequences of category values. This approach is inspired from neural sequence-to-sequence models commonly used for machine translation \citep{Cho2014a,Sutskever2014}.
For each word in a sentence, the 
decoder uses  a unidirectional LSTM network (Figure~\ref{fig:neural_models} (c)) to generate a sequence of morphological category-value pairs based on the context vector $h$  
and the previous predictions.
The probability of a morphological label $t$ is under this model:
\begin{equation}
p(t|h)_{\textsc{Seq}} = \prod_{j=0} \caret m p(f_j|f_0, \dots, f_{j-1}, h)
\end{equation}

Decoding starts by passing the start-of-sequence symbol  as input.
At each time step, the decoder computes the label context vector $g_j$ based on the previously predicted category value, previous label context vector and the word's context vector.
\begin{equation*}
g_j = \text{LSTM}([f_{j-1}; h], g_{j-1})
\end{equation*}
The probability of each morphological feature-value pair is then computed with a softmax.
\begin{equation*}
p(f_j|g_j)_{\textsc{Seq}} = \text{softmax}(W_{\textsc{seq}} g_j + b_{\textsc{seq}})
\end{equation*}
At training time, we feed correct labels as inputs while at inference time, we greedily emit the best prediction from the set of all possible feature-value pairs.
The decoding terminates once the end-of-sequence symbol is produced.


\subsection{Encoder}
\label{sec:encoder}
We adopt a standard sequence tagging encoder architecture for all our models. 
It consists of a  bidirectional LSTM network that maps words in a sentence into context vectors using character and word-level embeddings.
Character-level word embeddings are constructed with a bidirectional LSTM network and they capture useful information about words' morphology and shape.
Word level embeddings are initialised with pre-trained embeddings and fine-tuned during training.
The character and word-level embeddings are concatenated and passed as inputs to the bidirectional LSTM encoder.
The resulting hidden states $h_i$ capture contextual information for each word in a sentence.
Similar encoder architectures have been applied recently with notable success to morphological tagging \citep{heigold2017,yu2017} as well as several other sequence tagging tasks \citep{lample2016,Chiu2016,ling2015}.

\section{Experimental Setup}
\label{sec:experiments}

\begin{table*}
\centering
\ssmall
\renewcommand{\arraystretch}{0.98}
\tabcolsep=0.11cm
\begin{tabular}{lrrrrrr|rrrrr}
\toprule
 \multirow{3}{*}{\bf \footnotesize Dataset} & \multicolumn{6}{c|}{\bf \scriptsize Train set} & \multicolumn{5}{c}{\bf \scriptsize Test set} \\
 \cmidrule[\lightrulewidth](l){2-12}
&	\multirow{2}{*}{\bf Tokens} & \multirow{2}{*}{\bf Types}	&	\multicolumn{2}{c}{\bf Tags per word}	&	\multirow{2}{*}{\bf \% Emb}	&	\multirow{2}{*}{\bf \# Tags}	&	\multirow{2}{*}{\bf Tokens} &	\multirow{2}{*}{\bf Types}	& \multirow{2}{*}{\bf  \% OOV} &	\multicolumn{2}{c}{\bf OOV Tags} \\ 
 &	 &	&	\bf Avg & \bf Max	& & 	&	&	&	& \bf Tokens & \bf Types \\ 
\midrule
Afrikaans	&	33894	&	5080	&	1.1	&	4	&	62.7	&	61	&	10065	&	2476	&	13.8	&	3	&	3 \\ 
Arabic	&	254340	&	33225	&	1.8	&	10	&	90.1	&	349	&	32128	&	8754	&	9.8	&	6	&	6 \\ 
Basque	&	72974	&	19222	&	1.4	&	13	&	53.8	&	884	&	24374	&	8896	&	17.8	&	71	&	61 \\ 
Belarusian	&	5217	&	2303	&	1.4	&	6	&	74.6	&	346	&	1382	&	708	&	39.7	&	48	&	32 \\ 
Bulgarian	&	124336	&	25047	&	1.1	&	7	&	65.7	&	432	&	15724	&	5974	&	12.3	&	4	&	3 \\ 
Catalan	&	418494	&	31544	&	1.2	&	8	&	62.0	&	267	&	58017	&	9832	&	5.2	&	3	&	3 \\ 
Chinese	&	98608	&	17610	&	1.3	&	6	&	65.8	&	31	&	12012	&	4055	&	12.5	&	1	&	1 \\ 
Croatian	&	169283	&	34968	&	1.6	&	19	&	66.0	&	1105	&	13228	&	5513	&	14.1	&	13	&	13 \\ 
Czech	&	1175374	&	125358	&	1.7	&	25	&	59.7	&	2630	&	174252	&	37727	&	7.0	&	127	&	94 \\ 
Czech-CAC	&	473622	&	66272	&	1.7	&	21	&	72.4	&	1746	&	10900	&	4499	&	12.6	&	17	&	17 \\ 
Czech-CLTT	&	27005	&	4336	&	1.5	&	21	&	73.3	&	418	&	4126	&	1169	&	17.2	&	39	&	30 \\ 
Czech-FicTree	&	134059	&	25943	&	1.4	&	58	&	72.9	&	1464	&	16761	&	5691	&	12.8	&	46	&	43 \\ 
Danish	&	80378	&	16330	&	1.2	&	5	&	62.3	&	157	&	10023	&	3424	&	15.3	&	3	&	2 \\ 
Dutch	&	186046	&	26665	&	1.2	&	6	&	59.8	&	62	&	11046	&	3054	&	13.7	&	23	&	1 \\ 
Dutch-LassySmall	&	81243	&	14622	&	1.1	&	5	&	54.7	&	60	&	10080	&	3573	&	7.4	&	0	&	0 \\ 
English	&	204607	&	19672	&	1.4	&	10	&	58.3	&	117	&	25097	&	5630	&	9.1	&	3	&	3 \\ 
English-LinES	&	50095	&	7436	&	1.2	&	4	&	79.8	&	17	&	15623	&	3530	&	10.3	&	0	&	0 \\ 
English-ParTUT	&	43545	&	6963	&	1.3	&	8	&	74.7	&	133	&	3412	&	1136	&	9.3	&	3	&	3 \\ 
Estonian	&	85567	&	23055	&	1.3	&	7	&	58.0	&	662	&	10618	&	4928	&	18.6	&	28	&	24 \\ 
Finnish	&	162827	&	49210	&	1.1	&	9	&	59.4	&	2052	&	21070	&	9112	&	23.7	&	144	&	119 \\ 
Finnish-FTB	&	127845	&	39755	&	1.2	&	8	&	59.3	&	1762	&	16311	&	8011	&	23.0	&	83	&	76 \\ 
French	&	366371	&	42268	&	1.2	&	10	&	53.5	&	228	&	10298	&	3284	&	5.8	&	1	&	1 \\ 
French-ParTUT	&	24922	&	3815	&	1.3	&	10	&	87.3	&	197	&	2693	&	831	&	11.2	&	2	&	2 \\ 
French-Sequoia	&	51924	&	8463	&	1.2	&	5	&	73.2	&	200	&	10360	&	3023	&	8.9	&	0	&	0 \\ 
Galician	&	86676	&	13236	&	1.1	&	4	&	73.5	&	27	&	32390	&	7169	&	9.9	&	3	&	2 \\ 
Galician-TreeGal	&	5262	&	1873	&	1.3	&	9	&	77.7	&	173	&	10900	&	3182	&	26.8	&	81	&	41 \\ 
German	&	268145	&	49472	&	2.3	&	38	&	25.3	&	684	&	16537	&	5406	&	11.7	&	28	&	26 \\ 
Gothic	&	35024	&	6787	&	1.4	&	12	&	1.5	&	623	&	10182	&	2827	&	12.4	&	28	&	23 \\ 
Greek	&	43440	&	9049	&	1.3	&	15	&	74.4	&	349	&	10922	&	3370	&	16.4	&	9	&	6 \\ 
Hebrew	&	169360	&	29638	&	1.3	&	8	&	87.8	&	521	&	15134	&	5115	&	16.1	&	7	&	6 \\ 
Hindi	&	281057	&	16974	&	2.4	&	55	&	79.3	&	939	&	35430	&	5335	&	4.6	&	23	&	23 \\ 
Hungarian	&	20166	&	7767	&	1.4	&	5	&	75.7	&	580	&	10448	&	4558	&	37.1	&	108	&	85 \\ 
Indonesian	&	97531	&	19223	&	1.2	&	6	&	45.3	&	21	&	11780	&	4354	&	13.8	&	0	&	0 \\ 
Irish	&	3183	&	1257	&	1.5	&	8	&	62.3	&	236	&	10138	&	3245	&	36.1	&	276	&	113 \\ 
Italian	&	288750	&	28915	&	1.2	&	11	&	70.1	&	278	&	11153	&	3533	&	5.6	&	0	&	0 \\ 
Italian-ParTUT	&	52390	&	8323	&	1.1	&	6	&	82.0	&	205	&	3929	&	1318	&	9.1	&	1	&	1 \\ 
Italian-PoSTWITA	&	53725	&	12363	&	1.2	&	9	&	48.7	&	201	&	6778	&	2550	&	17.3	&	6	&	4 \\ 
Kazakh	&	547	&	343	&	1.2	&	2	&	73.2	&	72	&	10142	&	4559	&	71.9	&	2371	&	371 \\ 
Korean	&	52328	&	27714	&	1.1	&	4	&	68.8	&	11	&	10926	&	7060	&	37.5	&	0	&	0 \\ 
Latin	&	8018	&	3854	&	1.4	&	7	&	64.6	&	347	&	10954	&	4996	&	45.8	&	153	&	76 \\ 
Latin-ITTB	&	270403	&	12526	&	1.5	&	13	&	63.1	&	985	&	10561	&	1642	&	2.2	&	14	&	12 \\ 
Latin-PROIEL	&	147044	&	22258	&	1.4	&	21	&	50.6	&	993	&	12152	&	4331	&	9.8	&	15	&	13 \\ 
Latvian	&	62397	&	17745	&	1.3	&	30	&	64.0	&	742	&	14490	&	5467	&	23.9	&	46	&	36 \\ 
Lithuanian	&	3210	&	1522	&	1.2	&	3	&	73.2	&	297	&	1060	&	625	&	54.7	&	72	&	57 \\ 
Marathi	&	3253	&	969	&	1.6	&	70	&	78.1	&	261	&	448	&	199	&	26.3	&	19	&	15 \\ 
Norwegian-Bokmaal	&	243887	&	30072	&	1.2	&	6	&	61.8	&	203	&	29966	&	6616	&	11.3	&	4	&	3 \\ 
Norwegian-Nynorsk	&	245330	&	29133	&	1.3	&	8	&	50.0	&	184	&	24773	&	5963	&	11.1	&	3	&	2 \\ 
Old\_Church\_Slavonic	&	37432	&	7745	&	1.4	&	11	&	2.2	&	859	&	10031	&	3243	&	14.1	&	87	&	66 \\ 
Persian	&	122180	&	13859	&	1.1	&	5	&	89.7	&	162	&	16122	&	3945	&	8.5	&	3	&	2 \\ 
Polish	&	63070	&	21230	&	1.5	&	12	&	72.3	&	991	&	10906	&	5107	&	24.2	&	30	&	26 \\ 
Portuguese	&	222070	&	27396	&	1.4	&	35	&	61.9	&	375	&	10942	&	3417	&	8.2	&	3	&	3 \\ 
Portuguese-BR	&	273176	&	29944	&	1.2	&	8	&	58.5	&	22	&	33638	&	8047	&	6.8	&	0	&	0 \\ 
Romanian	&	185113	&	30970	&	1.2	&	6	&	69.3	&	451	&	16324	&	5755	&	10.4	&	7	&	6 \\ 
Russian	&	75964	&	25708	&	1.5	&	15	&	66.6	&	693	&	11548	&	5717	&	26.4	&	31	&	23 \\ 
Russian-SynTagRus	&	871082	&	107891	&	1.4	&	12	&	74.7	&	723	&	117470	&	29078	&	9.5	&	14	&	14 \\ 
Serbian	&	65764	&	14713	&	1.4	&	12	&	59.4	&	539	&	10891	&	4038	&	16.2	&	8	&	8 \\ 
Slovak	&	80575	&	21104	&	1.4	&	39	&	63.7	&	1199	&	13028	&	6049	&	35.8	&	72	&	58 \\ 
Slovenian	&	112530	&	29390	&	1.4	&	7	&	67.2	&	1101	&	14077	&	5856	&	19.9	&	20	&	19 \\ 
Slovenian-SST	&	9487	&	2672	&	1.4	&	5	&	90.5	&	500	&	10000	&	2812	&	21.6	&	202	&	132 \\ 
Spanish	&	389703	&	46979	&	1.4	&	12	&	56.7	&	399	&	12267	&	4114	&	7.4	&	3	&	3 \\ 
Spanish-AnCora	&	446145	&	38456	&	1.2	&	8	&	68.3	&	295	&	52801	&	10615	&	5.6	&	4	&	2 \\ 
Swedish	&	66645	&	12911	&	1.2	&	8	&	70.3	&	202	&	20377	&	5127	&	14.9	&	12	&	8 \\ 
Swedish-LinES	&	48325	&	9659	&	1.2	&	6	&	77.3	&	168	&	15029	&	4150	&	15.0	&	875	&	16 \\ 
Tamil	&	6849	&	3040	&	1.1	&	4	&	78.7	&	201	&	2183	&	1132	&	44.3	&	20	&	15 \\ 
Telugu	&	5082	&	1743	&	1.1	&	4	&	0.3	&	14	&	721	&	387	&	25.0	&	0	&	0 \\ 
Turkish	&	39169	&	14576	&	1.2	&	9	&	67.5	&	972	&	10256	&	5139	&	26.4	&	87	&	82 \\ 
Ukrainian	&	75054	&	23970	&	1.4	&	23	&	72.6	&	1197	&	14939	&	6337	&	27.2	&	72	&	60 \\ 
Urdu	&	108690	&	9547	&	2.7	&	52	&	73.9	&	1001	&	14806	&	2949	&	6.4	&	27	&	21 \\ 
Vietnamese	&	20285	&	3625	&	1.2	&	4	&	33.7	&	15	&	11955	&	2684	&	17.1	&	1	&	1 \\ 
\bottomrule
\end{tabular}

\caption{\small Descriptive statistics for all UDv2.1 datasets. For training sets we report the number of word tokens and types, the average (Avg) and maximum (Max) tags per word type, the proportion of word types for which pre-trained embeddings were available (\% Emb) and the size of the morphological tagset (\# Tags). For the test sets, we also give the total number of tokens and types, the proportion of OOV words (\% OOV) and the number of OOV tag tokens and types.}
\label{tbl:corpus_stats}
\end{table*}

This section details the experimental setup. We describe the data, then we introduce the baseline models and finally we report the hyperparameters of the models.

\subsection{Data}
We run experiments on the Universal Dependencies version 2.1~\citep{nivre2017}.
We excluded corpora that did not include train/dev/test split, word form information\footnote{French-FTB and Arabic-NYUAD}, or morphological features\footnote{Japanese}. Additionally, we excluded corpora for which pre-trained word embeddings were not available.\footnote{Ancient Greek and Coptic}
The resulting dataset contains 69 corpora covering 49 different languages.
Tagsets were constructed by concatenating the POS and morphological annotations of the treebanks.
Table~\ref{tbl:corpus_stats} gives corpus statistics. We present type and token counts for both training and test sets. For training set, we also show the average and maximum number of tags per word type and the size of the morphological tagset. 
For the test set, we report the proportion of out-of-vocabulary (OOV) words as well as the number of OOV tag tokens and types.

In the encoder, we use fastText word embeddings \citep{bojanowski2017} pre-trained on Wikipedia.\footnote{\ssmall \url{https://github.com/facebookresearch/fastText}}
Although these embeddings are uncased, our model still captures case information by means of character-level embeddings.
In Table~\ref{tbl:corpus_stats}, we also report for each language the proportion of word types for which the pre-trained embeddings are available.

\subsection{Baseline Models}

We use two models as baseline: the CRF-based \textsc{MarMoT} \citep{mueller2013} and the regular neural multiclass classifier.

\paragraph{MarMoT (\textsc{MMT})}
\textsc{MarMoT}\footnote{\url{http://cistern.cis.lmu.de/marmot/}} is a CRF-based morphological tagger which has been shown to achieve competitive performance across several languages \citep{mueller2013}.
\textsc{MarMoT} approximates the CRF objective using a pruning strategy which enables training higher-order models and handling large tagsets.
In particular, the tagger first predicts the POS part of the label and based on that, constrains the set of possible morphological labels.
Following the results of \citet{mueller2013}, we train second-order models. We tuned the regularization type and weight on German development set and based on that, we use L2 regularization with weight 0.01 in all our experiments.

\paragraph{Neural Multiclass classifier (\textsc{Mc})}
As the second baseline, we employ the standard multiclass classifier used by both \citet{heigold2017} and \citet{yu2017}.
The proposed model consists of an LSTM-based encoder, identical to the one described above in section~\ref{sec:encoder}, and a softmax classifier over the full tagset.
The tagset sizes for each corpora are shown in Table~\ref{tbl:corpus_stats}.
During preliminary experiments, we also added CRF layer on top of softmax, but as this made the decoding process considerably slower without any visible improvement in accuracy, we did not adopt CRF decoding here.
The multiclass model is shown in Figure~\ref{fig:neural_models} (d).

The inherent limitation of both baseline models is their inability to predict tags that are not present in the training corpus. Although the number of such tags in our data set is not large, it is nevertheless non-zero for most languages.

\subsection{Training and Parametrisation}\label{sbsec:training}
Since tuning model hyperparameters for each of the 69 datasets individually is computationally demanding, 
we optimise  parameters on Finnish---a morphologically complex language with a reasonable dataset size---and apply the resulting values to other languages.
We first tuned the character embedding size and character-LSTM hidden layer size of the encoder on the \textsc{Seq} model and reused the obtained values with all other models.
We tuned the batch size, the learning rate and the decay factor for the \textsc{Seq} and \textsc{Mc} models separately since these models are architecturally quite different.
For the \textsc{McMl} and \textsc{HMcMl} models we reuse the values obtained for the \textsc{Mc} model.
The remaining hyperparameter values are fixed.
Table~\ref{tbl:parameters} lists the hyperparameters for all models.

We train all neural models using stochastic gradient descent for up to 400 epochs and stop early if there has been no improvement on development set within 50 epochs.
For all models except \textsc{Seq}, we decay the learning rate by a factor of 0.98 after every 2500 batch updates.
We initialise biases with zeros and parameter matrices using Xavier uniform initialiser~\citep{Glorot2010}.

Words in training sets with no pre-trained embeddings are initialised with random embeddings.
At test time, words with no pre-trained embedding are assigned a special UNK-embedding.
We train the UNK-embedding by randomly substituting the singletons in a batch with the UNK-embedding with a probability of 0.5.

\begin{table}[t]
\centering
\footnotesize
\tabcolsep=0.11cm
\begin{tabular}{lrr}
\toprule
 & \textsc{Seq} & \textsc{Other NN} \\
\midrule
\textbf{Encoder} & &   \\
Word embedding size & 300 & 300 \\
Character embedding size & 100 & 100 \\ 
Character LSTM hidden layer size & 150 & 150  \\
Word embedding dropout & 0.5 & 0.5 \\
LSTM layers & 1 & 1 \\
LSTM hidden state size & 400 & 400 \\
LSTM input dropout & 0.5 & 0.5 \\
LSTM state dropout & 0.3 & 0.3 \\
LSTM output dropout & 0.5 & 0.5 \\
\midrule
\textbf{Decoder} & &   \\
LSTM hidden state size & 800 & 800 \\
Tag embedding size & 150 & -- \\
\midrule
\textbf{Training} & & \\
Initial learning rate & 1.0  & 1.0 \\
Batch size & 5 & 20 \\
Maximum epochs & 400 & 400 \\
Learning rate decay factor & -- & 0.98 \\
\bottomrule
\end{tabular}
\caption{Hyperparameters for neural models.}
\label{tbl:parameters}
\end{table}


\section{Results}
\label{sec:results}

\begin{table*}
\centering
\ssmall
\renewcommand{\arraystretch}{0.91}
\tabcolsep=0.095cm
\begin{tabular}{l >{\centering}m{0.7cm} >{\centering}m{0.7cm} >{\centering}m{0.7cm} >{\centering}m{0.9cm} r| >{\centering}m{0.7cm} >{\centering}m{0.7cm} >{\centering}m{0.7cm} >{\centering}m{0.7cm} r| >{\centering}m{0.7cm} >{\centering}m{0.7cm} >{\centering}m{0.7cm} >{\centering}m{0.7cm} r}
\toprule
& \multicolumn{5}{c|}{\bf Full tag (all words)} & \multicolumn{5}{c|}{\bf Full tag (OOV words)}  & \multicolumn{5}{c}{\bf POS (all words)} \\
\bf Dataset	&	\textsc{\bf MMT}	&	\textsc{\bf Mc}	&	\textsc{ \bf McMl}	&	\textsc{ \bf HMcMl}	&	\textsc{\bf  Seq}	&	\textsc{ \bf MMT}	&	\textsc{\bf Mc}	&	\textsc{\bf McMl}	&	\textsc{\bf HMcMl}	&	\textsc{\bf Seq}	&	\textsc{\bf MMT}	&	\textsc{\bf Mc}	&	\textsc{\bf McMl}	&	\textsc{\bf HMcMl}	&	\textsc{\bf Seq} \\
\midrule
\rowcolor{violet!30} Afrikaans	&	94.17	&	95.17	&	94.46	&	94.65	&	\textbf{95.45}	&	79.77	&	84.67	&	81.93	&	82.72	&	\textbf{84.88}	&	96.47	&	97.40	&	97.62	&	97.48	&	\textbf{97.66} \\ 
\rowcolor{pink!30} Arabic	&	90.96	&	93.39	&	93.25	&	93.23	&	\textbf{93.84}	&	81.25	&	86.06	&	85.24	&	85.14	&	\textbf{87.14}	&	95.22	&	96.01	&	96.18	&	96.20	&	\textbf{96.22} \\ 
\rowcolor{magenta!30} Basque	&	87.15	&	89.92	&	89.96	&	90.15	&	\textbf{90.33}	&	63.67	&	\textbf{72.65}	&	71.61	&	71.86	&	71.95	&	93.87	&	95.25	&	\textbf{96.00}	&	96.00	&	95.89 \\ 
\rowcolor{cyan!30} Belarusian	&	73.66	&	71.35	&	72.29	&	75.54	&	\textbf{78.15}	&	48.18	&	46.35	&	47.81	&	52.92	&	\textbf{59.12}	&	90.38	&	86.54	&	91.53	&	\textbf{93.42}	&	93.20 \\ 
\rowcolor{pink!30} Bulgarian	&	95.90	&	97.03	&	96.76	&	96.76	&	\textbf{97.04}	&	82.62	&	\textbf{88.74}	&	86.66	&	86.72	&	88.22	&	98.04	&	98.64	&	98.76	&	\textbf{98.82}	&	98.79 \\ 
\rowcolor{pink!30} Catalan	&	96.60	&	97.52	&	97.39	&	97.36	&	\textbf{97.59}	&	89.21	&	91.95	&	91.75	&	91.35	&	\textbf{92.28}	&	98.05	&	98.63	&	98.68	&	98.65	&	\textbf{98.70} \\ 
\rowcolor{magenta!30} Chinese	&	90.91	&	92.97	&	92.79	&	92.47	&	\textbf{93.27}	&	77.90	&	82.24	&	81.71	&	81.17	&	\textbf{82.91}	&	91.89	&	93.84	&	93.70	&	93.44	&	\textbf{94.11} \\ 
\rowcolor{pink!30} Croatian	&	84.99	&	88.66	&	88.96	&	88.96	&	\textbf{89.24}	&	66.11	&	74.87	&	75.89	&	\textbf{76.48}	&	76.37	&	96.47	&	97.25	&	\textbf{97.54}	&	97.41	&	97.45 \\ 
\rowcolor{pink!30} Czech	&	93.00	&	\textbf{95.81}	&	95.06	&	95.05	&	95.39	&	73.07	&	\textbf{82.92}	&	80.81	&	80.53	&	79.70	&	98.56	&	98.95	&	\textbf{99.00}	&	98.99	&	98.88 \\ 
\rowcolor{pink!30} Czech-CAC	&	90.46	&	\textbf{95.19}	&	94.74	&	94.72	&	95.14	&	69.39	&	\textbf{82.25}	&	80.13	&	79.91	&	81.59	&	98.65	&	99.06	&	99.17	&	\textbf{99.28}	&	99.05 \\ 
\rowcolor{violet!30} Czech-CLTT	&	89.21	&	89.63	&	90.45	&	91.01	&	\textbf{91.37}	&	73.00	&	77.78	&	78.48	&	78.20	&	\textbf{80.03}	&	98.01	&	97.99	&	98.91	&	\textbf{99.05}	&	98.67 \\ 
\rowcolor{pink!30} Czech-FicTree	&	91.24	&	93.93	&	94.54	&	94.48	&	\textbf{94.64}	&	75.32	&	83.96	&	84.48	&	83.87	&	\textbf{85.46}	&	97.55	&	98.14	&	\textbf{98.57}	&	98.51	&	98.38 \\ 
\rowcolor{magenta!30} Danish	&	93.90	&	95.73	&	95.26	&	95.46	&	\textbf{95.97}	&	78.74	&	85.24	&	83.03	&	83.68	&	\textbf{85.96}	&	95.79	&	97.26	&	97.30	&	97.44	&	\textbf{97.51} \\ 
\rowcolor{pink!30} Dutch	&	91.84	&	94.62	&	93.70	&	93.81	&	\textbf{94.73}	&	70.49	&	\textbf{81.23}	&	77.65	&	77.52	&	80.57	&	94.39	&	96.23	&	96.22	&	96.11	&	\textbf{96.35} \\ 
\rowcolor{magenta!30} Dutch-LassySmall	&	97.09	&	97.05	&	97.33	&	97.29	&	\textbf{97.54}	&	80.73	&	83.96	&	83.15	&	82.35	&	\textbf{84.10}	&	97.82	&	97.83	&	\textbf{98.41}	&	98.36	&	98.26 \\ 
\rowcolor{pink!30} English	&	93.03	&	\textbf{94.92}	&	94.40	&	94.36	&	94.80	&	76.22	&	\textbf{85.43}	&	83.33	&	83.38	&	84.69	&	94.54	&	\textbf{96.13}	&	96.09	&	95.96	&	96.06 \\ 
\rowcolor{magenta!30} English-LinES	&	95.03	&	\textbf{96.52}	&	96.36	&	96.39	&	96.36	&	83.72	&	\textbf{90.34}	&	89.41	&	90.09	&	89.23	&	95.03	&	\textbf{96.52}	&	96.36	&	96.39	&	96.36 \\ 
\rowcolor{violet!30} English-ParTUT	&	92.32	&	93.76	&	93.17	&	93.17	&	\textbf{94.17}	&	70.22	&	76.49	&	73.35	&	73.67	&	\textbf{81.82}	&	93.87	&	95.43	&	\textbf{96.10}	&	96.07	&	95.87 \\ 
\rowcolor{magenta!30} Estonian	&	91.40	&	93.28	&	93.17	&	93.25	&	\textbf{93.30}	&	79.25	&	84.78	&	84.42	&	84.32	&	\textbf{85.13}	&	95.54	&	96.61	&	96.74	&	\textbf{96.85}	&	96.68 \\ 
\rowcolor{pink!30} Finnish	&	91.41	&	93.13	&	93.18	&	93.29	&	\textbf{93.41}	&	78.35	&	84.05	&	\textbf{84.79}	&	84.71	&	84.71	&	95.68	&	96.55	&	97.02	&	\textbf{97.05}	&	96.79 \\ 
\rowcolor{pink!30} Finnish-FTB	&	90.59	&	93.91	&	\textbf{94.13}	&	93.88	&	91.93	&	76.06	&	84.65	&	\textbf{85.50}	&	85.24	&	80.85	&	93.36	&	95.73	&	\textbf{96.28}	&	96.19	&	94.56 \\ 
\rowcolor{pink!30} French	&	95.68	&	96.36	&	95.97	&	96.17	&	\textbf{96.39}	&	82.67	&	87.02	&	86.36	&	85.19	&	\textbf{87.85}	&	96.93	&	97.48	&	97.43	&	\textbf{97.50}	&	97.49 \\ 
\rowcolor{violet!30} French-ParTUT	&	92.91	&	93.50	&	93.28	&	92.94	&	\textbf{93.95}	&	71.10	&	\textbf{73.42}	&	70.10	&	70.10	&	72.43	&	95.77	&	96.10	&	\textbf{96.77}	&	96.73	&	96.77 \\ 
\rowcolor{magenta!30} French-Sequoia	&	95.99	&	96.66	&	96.51	&	96.31	&	\textbf{96.91}	&	76.99	&	\textbf{83.64}	&	80.82	&	80.39	&	82.23	&	97.68	&	98.06	&	\textbf{98.33}	&	98.17	&	98.32 \\ 
\rowcolor{magenta!30} Galician	&	96.97	&	97.65	&	97.72	&	97.70	&	\textbf{97.76}	&	84.94	&	88.66	&	88.98	&	88.85	&	\textbf{89.01}	&	97.10	&	97.80	&	97.87	&	97.84	&	\textbf{97.90} \\ 
\rowcolor{cyan!30} Galician-TreeGal	&	86.31	&	83.83	&	85.00	&	85.31	&	\textbf{86.61}	&	68.40	&	66.77	&	67.83	&	68.28	&	\textbf{71.80}	&	90.13	&	88.36	&	91.99	&	\textbf{92.00}	&	91.48 \\ 
\rowcolor{pink!30} German	&	80.81	&	87.98	&	87.11	&	87.16	&	\textbf{88.32}	&	63.12	&	\textbf{78.53}	&	75.00	&	76.14	&	78.37	&	92.60	&	94.47	&	94.56	&	\textbf{94.62}	&	94.35 \\ 
\rowcolor{violet!30} Gothic	&	87.09	&	86.49	&	86.25	&	86.86	&	\textbf{87.99}	&	\textbf{69.70}	&	65.59	&	60.84	&	62.03	&	65.27	&	95.47	&	94.48	&	95.45	&	\textbf{96.02}	&	95.59 \\ 
\rowcolor{violet!30} Greek	&	91.00	&	92.63	&	93.85	&	93.58	&	\textbf{94.14}	&	73.17	&	78.42	&	80.55	&	79.32	&	\textbf{81.89}	&	96.74	&	97.21	&	\textbf{97.80}	&	97.74	&	97.73 \\ 
\rowcolor{pink!30} Hebrew	&	93.19	&	95.05	&	94.73	&	94.60	&	\textbf{95.09}	&	81.05	&	87.90	&	86.87	&	86.63	&	\textbf{88.02}	&	96.15	&	\textbf{97.59}	&	97.59	&	97.53	&	97.56 \\ 
\rowcolor{pink!30} Hindi	&	89.00	&	\textbf{91.78}	&	91.47	&	91.34	&	91.75	&	62.35	&	\textbf{72.37}	&	69.99	&	68.77	&	71.70	&	96.20	&	97.00	&	\textbf{97.32}	&	97.22	&	97.03 \\ 
\rowcolor{violet!30} Hungarian	&	71.47	&	80.96	&	82.89	&	82.45	&	\textbf{84.12}	&	49.42	&	67.14	&	70.08	&	68.87	&	\textbf{72.01}	&	92.78	&	93.94	&	95.30	&	95.31	&	\textbf{95.44} \\ 
\rowcolor{magenta!30} Indonesian	&	93.56	&	\textbf{93.79}	&	93.73	&	93.74	&	93.65	&	88.22	&	88.04	&	\textbf{88.53}	&	88.16	&	87.67	&	93.57	&	93.81	&	93.81	&	\textbf{93.85}	&	93.69 \\ 
\rowcolor{cyan!30} Irish	&	\textbf{67.99}	&	60.73	&	62.02	&	61.95	&	65.81	&	\textbf{35.48}	&	28.05	&	29.50	&	28.70	&	34.50	&	83.62	&	79.10	&	84.01	&	\textbf{84.22}	&	83.63 \\ 
\rowcolor{pink!30} Italian	&	97.06	&	97.53	&	97.31	&	97.31	&	\textbf{97.61}	&	86.61	&	\textbf{88.87}	&	86.61	&	86.29	&	88.71	&	97.74	&	98.16	&	98.19	&	\textbf{98.32}	&	98.26 \\ 
\rowcolor{magenta!30} Italian-ParTUT	&	96.13	&	\textbf{97.12}	&	96.79	&	96.84	&	97.12	&	80.22	&	\textbf{90.81}	&	86.35	&	85.79	&	88.30	&	97.28	&	97.86	&	\textbf{98.14}	&	98.12	&	98.12 \\ 
\rowcolor{magenta!30} Italian-PoSTWITA	&	91.92	&	\textbf{93.79}	&	93.23	&	93.36	&	93.69	&	75.85	&	\textbf{82.34}	&	80.20	&	80.80	&	81.83	&	93.54	&	95.32	&	\textbf{95.72}	&	95.68	&	95.16 \\ 
\rowcolor{cyan!30} Kazakh	&	\textbf{37.19}	&	31.63	&	28.84	&	28.70	&	34.35	&	\textbf{20.97}	&	13.52	&	10.45	&	10.38	&	17.84	&	52.73	&	48.94	&	52.38	&	\textbf{54.74}	&	54.57 \\ 
\rowcolor{magenta!30} Korean	&	93.98	&	95.82	&	95.55	&	95.49	&	\textbf{95.87}	&	90.48	&	\textbf{93.51}	&	93.12	&	92.90	&	93.33	&	93.98	&	95.82	&	95.55	&	95.50	&	\textbf{95.87} \\ 
\rowcolor{cyan!30} Latin	&	64.94	&	64.10	&	65.35	&	65.88	&	\textbf{67.45}	&	41.05	&	42.54	&	42.58	&	43.30	&	\textbf{46.99}	&	80.73	&	80.97	&	84.84	&	\textbf{85.57}	&	84.81 \\ 
\rowcolor{pink!30} Latin-ITTB	&	92.98	&	95.18	&	\textbf{95.60}	&	95.57	&	95.27	&	68.26	&	74.78	&	\textbf{75.65}	&	74.35	&	72.61	&	97.30	&	98.12	&	98.30	&	\textbf{98.34}	&	98.17 \\ 
\rowcolor{pink!30} Latin-PROIEL	&	88.37	&	\textbf{90.64}	&	90.20	&	90.13	&	89.66	&	68.43	&	\textbf{78.39}	&	74.46	&	73.20	&	71.69	&	95.78	&	96.68	&	\textbf{96.80}	&	96.72	&	95.94 \\ 
\rowcolor{magenta!30} Latvian	&	85.59	&	87.67	&	87.14	&	87.14	&	\textbf{87.79}	&	67.91	&	73.59	&	71.94	&	71.94	&	\textbf{73.88}	&	92.80	&	94.38	&	94.87	&	\textbf{94.88}	&	94.55 \\ 
\rowcolor{cyan!30} Lithuanian	&	65.00	&	58.02	&	64.91	&	63.58	&	\textbf{67.92}	&	44.66	&	36.72	&	43.79	&	43.10	&	\textbf{51.03}	&	73.87	&	70.00	&	81.60	&	79.25	&	\textbf{81.70} \\ 
\rowcolor{cyan!30} Marathi	&	66.07	&	68.75	&	64.06	&	64.96	&	\textbf{70.09}	&	39.83	&	\textbf{49.15}	&	33.05	&	36.44	&	44.92	&	82.14	&	82.81	&	84.15	&	\textbf{84.82}	&	84.60 \\ 
\rowcolor{pink!30} Norwegian-Bokmaal	&	94.99	&	96.37	&	96.13	&	95.94	&	\textbf{96.54}	&	80.14	&	84.53	&	83.11	&	82.54	&	\textbf{84.68}	&	97.33	&	98.24	&	98.39	&	98.26	&	\textbf{98.44} \\ 
\rowcolor{pink!30} Norwegian-Nynorsk	&	94.65	&	\textbf{96.25}	&	95.69	&	95.69	&	96.07	&	81.32	&	\textbf{85.30}	&	81.93	&	82.11	&	83.82	&	97.08	&	98.12	&	\textbf{98.22}	&	98.14	&	98.08 \\ 
\rowcolor{violet!30} Old\_Church\_Slavonic	&	87.58	&	86.96	&	87.01	&	86.87	&	\textbf{87.96}	&	60.31	&	\textbf{60.59}	&	57.49	&	57.13	&	58.83	&	94.98	&	94.40	&	95.38	&	\textbf{95.61}	&	94.94 \\ 
\rowcolor{pink!30} Persian	&	95.84	&	96.75	&	96.38	&	96.38	&	\textbf{96.79}	&	79.36	&	\textbf{86.09}	&	84.04	&	83.67	&	85.43	&	96.39	&	97.13	&	97.11	&	97.10	&	\textbf{97.30} \\ 
\rowcolor{magenta!30} Polish	&	86.04	&	90.46	&	\textbf{90.99}	&	90.78	&	90.99	&	69.13	&	\textbf{81.21}	&	79.36	&	79.81	&	80.87	&	96.65	&	97.73	&	\textbf{98.25}	&	98.11	&	98.04 \\ 
\rowcolor{pink!30} Portuguese	&	94.21	&	95.59	&	95.34	&	95.59	&	\textbf{95.75}	&	79.48	&	86.66	&	86.66	&	\textbf{86.77}	&	86.43	&	97.21	&	97.72	&	\textbf{98.06}	&	97.95	&	98.04 \\ 
\rowcolor{pink!30} Portuguese-BR	&	97.59	&	98.20	&	98.20	&	98.14	&	\textbf{98.21}	&	92.30	&	95.20	&	\textbf{95.56}	&	95.03	&	95.16	&	97.60	&	98.20	&	98.21	&	98.16	&	\textbf{98.22} \\ 
\rowcolor{pink!30} Romanian	&	96.30	&	97.00	&	96.72	&	96.61	&	\textbf{97.16}	&	85.15	&	89.51	&	88.10	&	87.92	&	\textbf{89.75}	&	97.18	&	97.61	&	97.74	&	\textbf{97.78}	&	97.77 \\ 
\rowcolor{magenta!30} Russian	&	85.99	&	90.21	&	90.73	&	90.93	&	\textbf{91.05}	&	66.91	&	77.85	&	78.24	&	78.90	&	\textbf{79.26}	&	95.42	&	96.43	&	96.72	&	\textbf{96.84}	&	96.50 \\ 
\rowcolor{pink!30} Russian-SynTagRus	&	94.44	&	\textbf{96.78}	&	96.48	&	96.58	&	96.67	&	78.91	&	\textbf{88.50}	&	87.21	&	87.48	&	86.98	&	98.51	&	98.84	&	98.92	&	98.93	&	\textbf{98.94} \\ 
\rowcolor{magenta!30} Serbian	&	91.17	&	93.25	&	93.32	&	93.58	&	\textbf{93.93}	&	77.32	&	83.22	&	82.20	&	82.48	&	\textbf{83.50}	&	97.47	&	97.89	&	\textbf{98.25}	&	98.17	&	98.19 \\ 
\rowcolor{magenta!30} Slovak	&	81.72	&	87.50	&	88.16	&	\textbf{88.54}	&	88.46	&	68.42	&	78.66	&	78.98	&	79.24	&	\textbf{79.69}	&	94.62	&	95.85	&	\textbf{96.49}	&	96.34	&	96.46 \\ 
\rowcolor{pink!30} Slovenian	&	89.39	&	94.32	&	94.05	&	93.98	&	\textbf{94.62}	&	73.14	&	86.34	&	83.94	&	83.58	&	\textbf{86.41}	&	97.07	&	98.15	&	98.29	&	98.39	&	\textbf{98.42} \\ 
\rowcolor{cyan!30} Slovenian-SST	&	78.71	&	75.75	&	79.18	&	80.02	&	\textbf{80.44}	&	45.45	&	44.06	&	48.40	&	49.88	&	\textbf{52.24}	&	88.44	&	87.54	&	92.04	&	\textbf{92.38}	&	90.99 \\ 
\rowcolor{pink!30} Spanish	&	94.33	&	\textbf{95.05}	&	94.82	&	94.81	&	94.90	&	77.34	&	\textbf{82.95}	&	82.29	&	82.18	&	81.52	&	95.88	&	96.89	&	96.95	&	\textbf{96.98}	&	96.83 \\ 
\rowcolor{pink!30} Spanish-AnCora	&	97.13	&	\textbf{97.67}	&	97.54	&	97.58	&	97.63	&	90.26	&	93.22	&	93.09	&	93.19	&	\textbf{93.36}	&	98.25	&	98.64	&	98.75	&	\textbf{98.78}	&	98.68 \\ 
\rowcolor{magenta!30} Swedish	&	94.28	&	95.41	&	95.07	&	95.25	&	\textbf{95.65}	&	82.72	&	86.11	&	84.20	&	84.07	&	\textbf{86.37}	&	96.38	&	97.49	&	97.69	&	\textbf{97.72}	&	97.66 \\ 
\rowcolor{violet!30} Swedish-LinES	&	85.24	&	86.38	&	85.99	&	85.98	&	\textbf{86.47}	&	64.01	&	\textbf{68.33}	&	66.28	&	65.79	&	67.26	&	95.00	&	96.17	&	\textbf{96.69}	&	96.65	&	96.25 \\ 
\rowcolor{cyan!30} Tamil	&	81.40	&	82.18	&	83.05	&	81.26	&	\textbf{85.75}	&	67.87	&	71.90	&	72.42	&	70.56	&	\textbf{75.83}	&	86.39	&	87.49	&	\textbf{91.07}	&	90.24	&	90.75 \\ 
\rowcolor{cyan!30} Telugu	&	\textbf{92.23}	&	90.43	&	89.04	&	89.32	&	91.26	&	\textbf{80.00}	&	75.56	&	70.00	&	71.67	&	78.33	&	\textbf{92.23}	&	90.43	&	89.04	&	89.32	&	91.26 \\ 
\rowcolor{violet!30} Turkish	&	86.09	&	89.47	&	90.69	&	90.51	&	\textbf{90.70}	&	63.97	&	74.85	&	\textbf{79.83}	&	79.02	&	79.13	&	92.86	&	94.67	&	\textbf{95.54}	&	95.51	&	95.19 \\ 
\rowcolor{magenta!30} Ukrainian	&	85.33	&	88.98	&	89.94	&	\textbf{89.96}	&	89.81	&	69.19	&	78.89	&	79.24	&	79.34	&	\textbf{79.36}	&	95.97	&	96.40	&	\textbf{97.23}	&	97.06	&	97.03 \\ 
\rowcolor{pink!30} Urdu	&	77.37	&	80.09	&	79.52	&	78.54	&	\textbf{80.66}	&	54.99	&	64.54	&	60.30	&	61.68	&	\textbf{65.07}	&	92.56	&	93.29	&	\textbf{93.87}	&	93.71	&	93.81 \\ 
\rowcolor{violet!30} Vietnamese	&	86.13	&	\textbf{88.66}	&	88.51	&	88.22	&	88.44	&	55.19	&	\textbf{70.81}	&	70.46	&	69.29	&	68.70	&	86.15	&	\textbf{88.67}	&	88.58	&	88.34	&	88.46 \\ 
\midrule
\rowcolor{pink!30} Average (>100K) & 92.18 & \textbf{94.37} & 94.12 & 94.07 & \textbf{94.37} & 76.65 & \textbf{84.03} & 82.67 & 82.47 & 83.42 & 96.49 & 97.37 & \textbf{97.52} & 97.50 & 97.40 \\
\rowcolor{magenta!30} Average (50K-100K) & 91.27 & 93.36 & 93.36 & 93.40 & \textbf{93.66} & 76.96 & \textbf{83.46} & 82.39 & 82.43 & 83.40 & 95.39 & 96.43 & \textbf{96.71} & 96.67 & 96.65 \\
\rowcolor{violet!30} Average (20K-50K)  & 87.56 & 89.42 & 89.69 & 89.66 & \textbf{90.43} & 66.35 & 72.55 & 71.76 & 71.47 & \textbf{73.84} & 94.37 & 95.13 & 95.83 & \textbf{95.86} & 95.69 \\
\rowcolor{cyan!30} Average (<20K) & 71.35 & 68.68 & 69.37 & 69.65 & \textbf{72.78} & 49.19 & 47.46 & 46.58 & 47.52 & \textbf{53.26} & 82.07 & 80.22 & 84.27 & 84.60 & \textbf{84.70} \\
\midrule
Overall average & 88.18 & 89.58 & 89.61 & 89.64 & \textbf{90.42} & 71.12 & 76.74 & 75.62 & 75.64 & \textbf{77.52} & 93.76 & 94.27 & 95.11 & \textbf{95.14} & 95.08 \\
\bottomrule
\end{tabular}

\caption{\small Morphological tagging accuracies on UDv2.1 test sets for MarMot (\textsc{MMT})  and \textsc{Mc} baselines as well as for \textsc{McMl}, \textsc{HMcMl} and \textsc{Seq} compositional models. The left section shows the full \textsc{Pos+Morph} tag results, the middle section gives accuracies for OOV words only, the right-most section shows the POS tagging accuracy. The best result in each section for each language is in bold. The languages are color-coded according to the training set size, lighter color denotes larger training set: cyan (<20K), violet (20K-50K), magenta (50K-100K), pink (>100K).}
\label{tbl:results}
\end{table*}

Table~\ref{tbl:results} presents the experimental results. We report tagging accuracy for all word tokens and also for OOV tokens only. 
A full morphological tag is considered correct if both its POS and all morphological features are correctly predicted. 

First of all, we can confirm the results of \citet{heigold2017} that the performance of neural morphological tagging indeed exceeds the results of a CRF-based model. In fact, all our neural models perform significantly better than \textsc{MarMoT} ($p<0.001$).\footnote{As indicated by Wilcoxon signed-rank test.}


The best neural model on average is the \textsc{Seq} model, which is significantly better from both the \textsc{Mc} baseline as well as the other two compositional models, whereby the improvement is especially well-pronounced on smaller datasets. We do not observe any significant differences between \textsc{McMl} and \textsc{HMcMl} models neither on all words nor OOV evaluation setting.

We also present POS tagging results in the right-most section of Table~\ref{tbl:results}. Here again, all neural models are better than CRF which is in line with the results presented by \citet{plank2016}. For POS tags, the \textsc{HMcMl} is the best on average. It is also significantly better than the neural \textsc{Mc} baseline, however, the differences with the \textsc{McMl} and \textsc{Seq} models are insignificant.

In addition to full-tag accuracies, we assess the performance on individual features.
Table~\ref{tbl:features} reports macro-averaged F1-cores for the \textsc{Seq} and the \textsc{Mc} models on universal features.
Results indicate that the \textsc{Seq} model systematically outperforms the \textsc{Mc} model on most features.

\begin{table}[t]
\centering
\small
\renewcommand{\arraystretch}{0.91}
\tabcolsep=0.095cm
\begin{tabular}{lrrr|lrrr}
\toprule
Feature & \textsc{Seq} & \textsc{Mc} & \# & Feature & \textsc{Seq} & \textsc{Mc} & \# \\
\midrule
POS & \textbf{91.03} & 90.20 & 69 & NumType & \textbf{89.68} & 87.82 & 54 \\
Number & \textbf{94.02} & 93.05 & 63 & Polarity & \textbf{93.83} & 92.86 & 54 \\
VerbForm & \textbf{91.29} & 89.86 & 61 & Degree & \textbf{87.44} & 84.12 & 48 \\
Person & \textbf{89.02} & 87.52 & 60 & Poss & \textbf{94.52} & 93.60 & 44 \\
Tense & \textbf{92.96} & 91.31 & 59 & Voice & \textbf{88.40} & 82.85 & 42 \\
PronType & \textbf{89.83} & 88.81 & 58 & Definite & \textbf{95.26} & 94.10 & 37 \\
Mood & \textbf{87.34} & 85.40 & 58 & Aspect & \textbf{89.76} & 87.71 & 29 \\
Gender & \textbf{89.31} & 87.78 & 55 & Animacy & \textbf{86.22} & 83.73 & 19 \\
Case & \textbf{88.90} & 87.04 & 55 & Polite & 75.76 & \textbf{80.48} & 10 \\
\bottomrule
\end{tabular}
\caption{Performance of \textsc{Seq} and \textsc{Mc} models on individual features reported as macro-averaged F1-scores. 
}
\label{tbl:features}
\end{table}



\section{Analysis and Discussion}
\label{sec:discussion}


\paragraph{OOV label accuracy}

\begin{figure}
\centering
\includegraphics[width=0.95\columnwidth]{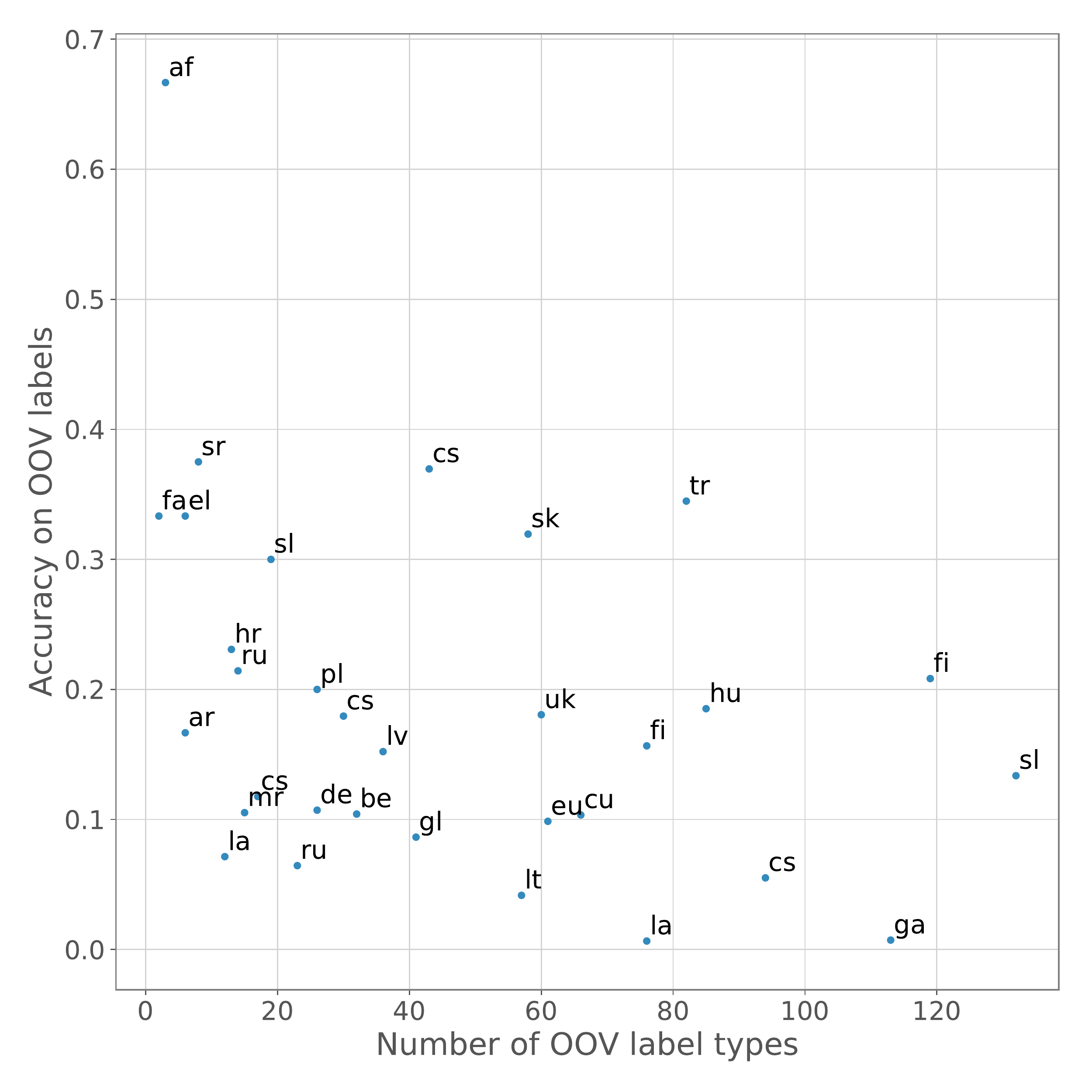}
\caption{OOV label accuracies of the \textsc{Seq} model.}
\label{fig:oov_labels_seq}
\end{figure}

\begin{figure}[ht]
\centering
\includegraphics[width=\columnwidth]{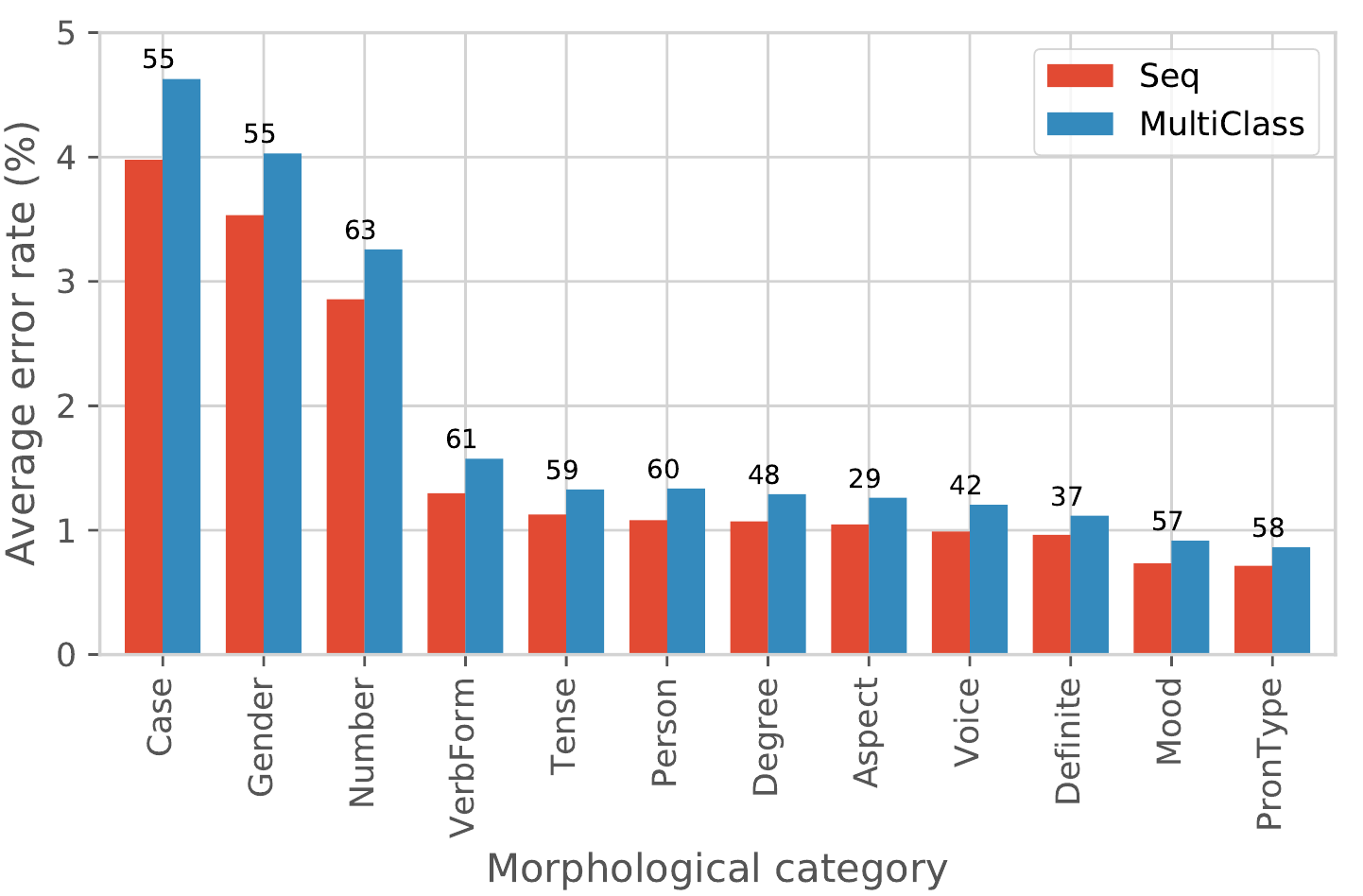}
\caption{Average error rates of distinct morphological categories for \textsc{Seq} and \textsc{Mc} models.}
\label{fig:cat_errors}
\end{figure}

Our models are able to predict labels that were not seen in the training data. Figure~\ref{fig:oov_labels_seq} presents the accuracy of test tokens with OOV labels obtained with our best performing \textsc{Seq} model plotted against the number of OOV label types. The datasets with zero accuracy are omitted. The main observation is that although the OOV label accuracy is zero for some languages, it is above zero on ca. half of the datasets---a result that would be impossible with \textsc{MarMoT} or \textsc{Mc} baselines. 






\paragraph{Error Analysis}
Figure~\ref{fig:cat_errors} shows the largest error rates for distinct morphological categories for both \textsc{Seq} and \textsc{Mc} models averaged over all languages. 
We observe that the error patterns are similar for both models but the error rates of the \textsc{Seq} model are consistently lower as expected. 

\paragraph{Stability Analysis}
To assess the stability of our predictions, we picked five languages from different families and with different corpus size, and performed five independent train/test runs for each language.
Table~\ref{tbl:stability} summarises the results of these experiments and demonstrates a reasonably small variance for all languages. 
For all languages, except for Finnish, the worst accuracy of the \textsc{Seq} model was better than the best accuracy of the \textsc{Mc} model, confirming our results that in those languages, the \textsc{Seq} model is consistently better than the \textsc{Mc} baseline. 

\begin{table}[t]
\centering
\small
\begin{tabular}{lcc}
\toprule
Dataset         & \textsc{Seq}     & \textsc{Mc} \\ 
\midrule
Finnish     & 93.24 $\pm$ 0.12 &  93.20 $\pm$ 0.07 \\ 
German      & 88.45 $\pm$ 0.21 &  87.74 $\pm$ 0.17 \\ 
Hungarian   & 84.51 $\pm$ 0.54 &  80.68 $\pm$ 0.48 \\ 
Russian     & 91.08 $\pm$ 0.18 &  90.13 $\pm$ 0.15 \\ 
Turkish     & 90.29 $\pm$ 0.24 &  89.16 $\pm$ 0.27 \\ 
\bottomrule
\end{tabular}
\caption{Mean accuracy with standard deviation over five independent runs for \textsc{Seq} and \textsc{Mc} models.}
\label{tbl:stability}
\vspace{-1em}
\end{table}

\paragraph{Hyperparameter Tuning}
It is possible that the hyperparameters tuned on Finnish are not optimal for other languages and thus, tuning hyperparameters for each language individually would lead to different conclusions than currently drawn.
To shed some light on this issue, we tuned hyperparameters for the \textsc{Seq} and \textsc{Mc} models on the same subset of five languages.
We first independently optimised the dropout rates on word embeddings, encoder's LSTM inputs and outputs, as well as the number of LSTM layers.
We then performed a grid search to find the optimal initial learning rate, the learning rate decay factor and the decay step.
Value ranges for the tuned parameters are given in Table~\ref{tbl:tuning_grid}.

\begin{table}[h]
\centering
\small
\begin{tabular}{lc}
\toprule
Parameter         & Values \\
\midrule
Word embedding dropout      & $\{0, 0.1, \dots, 0.5\}$ \\
LSTM input dropout          & $\{0, 0.1, \dots, 0.5\}$ \\
LSTM input dropout          & $\{0, 0.1, \dots, 0.5\}$ \\
Number of LSTM layers       & $\{1, 2\}$ \\
\midrule
Initial learning rate       & $\{0.01, 0.1, 1, 2\}$ \\
Learning rate decay factor  & $\{0.97, 0.98, 0.99, 1\}$ \\
Decay step                  & $\{1250, 2500, 5000\}$ \\
\bottomrule
\end{tabular}
\caption{The grid values for hyperparameter tuning.}
\label{tbl:tuning_grid}
\vspace{-1em}
\end{table}

Table~\ref{tbl:tuning} reports accuracies for the tuned models compared to the mean accuracies reported in Table~\ref{tbl:stability}.
As expected, both tuned models demonstrate superior performance on all languages, except for German with the \textsc{Seq} model.
Hyperparameter tuning has a greater overall effect on the \textsc{Mc} model, which suggests that it is more sensitive to the choice of parameters than the \textsc{Seq} model.
Still, the tuned \textsc{Seq} model performs better or at least as good as the \textsc{Mc} model on all languages.

\begin{table}[t]
\centering
\small
\begin{tabular}{lcc|cc}
\toprule
Dataset     & \textsc{Seq} & Gain  & \textsc{Mc} & Gain \\ 
\midrule
Finnish     & 93.44 & $+0.20$ & 93.43 & $+0.23$ \\
German      & 88.35 & $-0.10$ & 88.14 & $+0.40$ \\
Hungarian   & 85.56 & $+1.05$ & 82.29 & $+1.61$ \\
Russian     & 91.44 & $+0.36$ & 90.74 & $+0.61$ \\
Turkish     & 90.56 & $+0.27$ & 89.32 & $+0.16$ \\
\bottomrule
\end{tabular}
\caption{Accuracies of the tuned \textsc{Seq} and \textsc{Mc} models compared to the mean accuracies in Table~\ref{tbl:stability}.}
\label{tbl:tuning}
\end{table}

\paragraph{Comparison with Previous Work}
Since UD datasets have been in rapid development and different UD versions do not match, direct comparison of our results to previously published results is difficult.
Still, we show the results taken from \citet{heigold2017}, which were obtained on UDv1.3, to provide a very rough comparison.
In addition, we compare our \textsc{Seq} model with a neural tagger presented by \citet{Dozat2017}, which is similar to our \textsc{Mc} model, but employs a more sophisticated encoder.
We train this model on UDv2.1 on the same set of languages used by \citet{heigold2017}.

Table~\ref{tbl:result_comparison} reports evaluation results for the three models.
The \textsc{Seq} model and Dozat's tagger demonstrate comparable performance.
This suggests that the \textsc{Seq} model can be further improved by adopting a more advanced encoder from \citet{Dozat2017}.

\begin{table}
\centering
\small
\begin{tabular}{lcc|c}
\toprule
Dataset & \textsc{Seq} & Dozat & Heigold \\
\midrule
Arabic & \textbf{93.84} & 92.85 & 93.78 \\
Bulgarian & 97.04 & \textbf{97.25} & 95.14 \\
Czech & \bf 95.39 & 95.22 & 96.32 \\
English & 94.80 & \textbf{94.81} & 93.32 \\
Estonian & 93.30 & \bf 93.90 & 94.25 \\
Finnish & 93.41 & \textbf{93.73} & 93.52 \\
French & \textbf{96.39} & 95.90 & 94.91 \\
Hindi & 91.75 & \textbf{92.36}  & 90.84 \\
Hungarian & \textbf{84.12} & 82.84 & 77.59 \\
Romanian & 97.16 & \textbf{97.20} & 94.12 \\
Russian-SynTagRus & \textbf{96.67} & 96.20 & 96.45 \\
Turkish & \textbf{90.70} & 90.22 & 89.12 \\
\midrule
Average & \bf 93.71 & 93.54 & 92.45 \\
\bottomrule
\end{tabular}
\caption{Accuracies for the \textsc{SEQ} model, \citet{Dozat2017} and \citet{heigold2017}.}
\label{tbl:result_comparison}
\vspace{-1em}
\end{table}

\section{Conclusion}
\label{sec:conclusion}
\vspace{-0.4em}
We hypothesised that explicitly modeling the internal structure of complex labels for morphological tagging improves the overall tagging accuracy over the baseline with monolithic tags.
To test this hypothesis, we experimented with three approaches to model composite morphological tags in a neural sequence tagging framework.
Experimental results on 49 languages demonstrated the advantage of modeling morphological labels as sequences of category values, whereas the superiority of this model is especially pronounced on smaller datasets.
Furthermore, we showed that, in contrast to baselines, our models are capable of predicting labels that were not seen during training.

\section*{Acknowledgments}
This work was supported by the Estonian Research Council (grants no. 2056, 1226 and IUT34-4).

\bibliography{paper}
\bibliographystyle{acl_natbib_nourl}

\end{document}